\documentclass[sigconf]{acmart}

\AtBeginDocument{%
  }

\copyrightyear{2027}
\acmYear{2027}
\setcopyright{cc}
\setcctype{by}
\acmConference[WSDM '27]{Proceedings of the Twentieth ACM International Conference on Web Search and Data Mining}{February 15--19, 2027}{Hong Kong, Hong Kong}
\acmBooktitle{Proceedings of the Twentieth ACM International Conference on Web Search and Data Mining (WSDM '27), February 15--19, 2027, Hong Kong, Hong Kong}
\acmDOI{10.1145/3862322.3863553}
\acmISBN{979-8-4007-3117-4/2027/02}

\usepackage{amsmath,amsfonts}
\ifdefined\symbfit
  \providecommand{\bm}[1]{\symbfit{#1}}
\else
  \usepackage{bm}
\fi

\def\eqref#1{equation~\ref{#1}}

\def\1{\bm{1}}

\def\vc{{\bm{c}}}

\def\vp{{\bm{p}}}

\def\mC{{\bm{C}}}

\def\mR{{\bm{R}}}

\def\mX{{\bm{X}}}
\def\mY{{\bm{Y}}}

\ifdefined\mathsfit\else
\DeclareMathAlphabet{\mathsfit}{\encodingdefault}{\sfdefault}{m}{sl}
\SetMathAlphabet{\mathsfit}{bold}{\encodingdefault}{\sfdefault}{bx}{n}
\fi

\def\gF{{\mathcal{F}}}

\def\sR{{\mathbb{R}}}

\usepackage{color}
\usepackage{kotex}              
\usepackage{enumitem}           
\usepackage{subcaption}
\usepackage{multicol}
\usepackage{multirow}
\usepackage{makecell}

\usepackage[normalem]{ulem}
\useunder{\uline}{\ul}{}
\newcommand{\std}[1]{\scriptsize{$\pm$#1}}
\usepackage{float}
\usepackage{etoolbox}

\begin{document}


\title{Pseudo-Label-Triggered Retraining from Forecast Errors for Online Time Series Forecasting}

\author{Yeryeong Kwak}
\authornote{Equal contribution}
\email{yeroong9698@snu.ac.kr}
\orcid{0009-0007-3656-5221}
\affiliation{%
  \institution{Seoul National University}
  \city{Seoul}
  \country{Republic of Korea}
}

\author{Yoo-Min Jung}
\authornotemark[1]
\email{pamela7384@gmail.com}
\orcid{0000-0002-3352-5714}
\affiliation{%
  \institution{Seoul National University}
  \city{Seoul}
  \country{Republic of Korea}
}

\author{Jonghun Park}
\email{jonghun@snu.ac.kr}
\orcid{0000-0001-7505-110X}
\affiliation{%
  \institution{Seoul National University}
  \city{Seoul}
  \country{Republic of Korea}
}


\begin{abstract}
Real-world time series forecasting systems operate under non-stationary data streams, where forecasting performance may degrade over time.
Although retraining can recover the performance, it incurs non-trivial computational and operational costs.
Under limited deployment resources, the key challenge is therefore not only how to retrain but also when to retrain.
While existing retraining policies often rely on indirect indicators such as drift alarms or model staleness, we instead use realized forecast errors as direct deployment feedback.
In this paper, we propose \textbf{PILOT} (\textbf{P}seudo-label-\textbf{I}nformed \textbf{L}earned \textbf{O}nline \textbf{T}rigger), an online retraining framework that learns when to retrain from forecast-error dynamics.
Since ground-truth retraining labels are unavailable, PILOT constructs a pseudo-label from future increases in forecast error and trains a lightweight scorer to predict it from observed error states.
At deployment, PILOT uses only completed forecast errors and serves as a plug-in module for arbitrary forecasting backbones without architectural modification.
We evaluate PILOT under standard multivariate forecasting settings across eight benchmarks with three representative backbones---DLinear, iTransformer, and TimesNet.
Across all three backbones, PILOT achieves state-of-the-art average-rank performance among retraining policies while maintaining a favorable performance--efficiency trade-off.
\end{abstract}

\ccsdesc[500]{Computing methodologies~Online learning settings}
\ccsdesc[500]{Applied computing~Forecasting}
\ccsdesc[500]{Mathematics of computing~Time series analysis}

\keywords{time series forecasting, pseudo-labeling, model retraining, online learning, concept drift}


\maketitle


\section{Introduction}
\label{sec:introduction}
Real-world time series forecasting models operate in online settings~\cite{hoi2021online} and face distribution shifts~\cite{liu2023handling,lu2019conceptdrift} that can progressively degrade forecasting performance~\cite{zhang2024d3a}. 
Retraining on recent observations can restore performance~\cite{liu2023handling,mahadevan2024cara,zanotti2025retrainfreq}. 
However, retraining incurs non-trivial computational overhead~\cite{mahadevan2024cara,zanotti2025retrainfreq,hoffman2024useful}, may disrupt deployment pipelines~\cite{derakhshan2019continuous}, and does not necessarily improve forecasting performance~\cite{regol2025upf,zanotti2025retrainfreq}. 
Under resource constraints, retraining should therefore be invoked selectively rather than routinely~\cite{zliobaite2015costsensitive,mahadevan2024cara,regol2025upf,hoffman2024useful}.

The key question is therefore not only how to retrain, but also when to retrain. 
Existing policies trigger updates based on predicted future performance~\cite{regol2025upf}, model staleness or expected return~\cite{mahadevan2024cara,zliobaite2015costsensitive}, or statistical changes in the data stream~\cite{bifet2007adwin,raab2020kswin}. 
However, these criteria do not directly use realized forecasting performance as deployment feedback.

In contrast, forecasting systems naturally produce direct deployment feedback once target observations become available: forecast errors.
Prior work has primarily used them for monitoring~\cite{trigg1964monitoring,grundy2026online}. 
We instead use recent error dynamics to anticipate performance degradation and selectively allocate retraining computation.

In this paper, we propose \textbf{PILOT} (\textbf{P}seudo-label-\textbf{I}nformed \textbf{L}earned \textbf{O}nline \textbf{T}rigger), a lightweight framework that learns retraining decisions from forecast-error dynamics. 
During offline training, PILOT constructs a pseudo-label from future forecast-error increase relative to a recent baseline and trains a compact scorer to predict this signal from recent error states.
This supervision requires no counterfactual retraining simulation. 
At deployment, the scorer triggers retraining using only realized forecast errors.
Since it relies only on forecast errors, PILOT can be attached to different forecasting backbones without architectural modification.

Our contributions are as follows:
\begin{itemize}[leftmargin=15pt, topsep=1pt, itemsep=1pt]
    \item We formulate retraining timing in online forecasting as a decision problem driven by realized forecast-error dynamics.
    \item To the best of our knowledge, PILOT is the first approach in online time series forecasting to use pseudo-labeling for retraining decisions, enabling the decision module to be trained without ground-truth retraining labels.
    \item Unlike representative retraining studies using synthetic or custom-designed streams~\cite{mahadevan2024cara,regol2025upf}, we evaluate PILOT on eight widely adopted multivariate forecasting benchmarks.
    \item Across three representative backbones spanning linear, Transformer, and convolutional architectures~\cite{zeng2023dlinear,liu2024itransformer,wu2023timesnet}, PILOT achieves state-of-the-art performance among retraining policies with a favorable performance--efficiency trade-off.
\end{itemize}

\begin{figure*}[!t]
  \centering
  \begin{subfigure}[b]{0.85\textwidth}
    \centering
    \includegraphics[width=\textwidth]{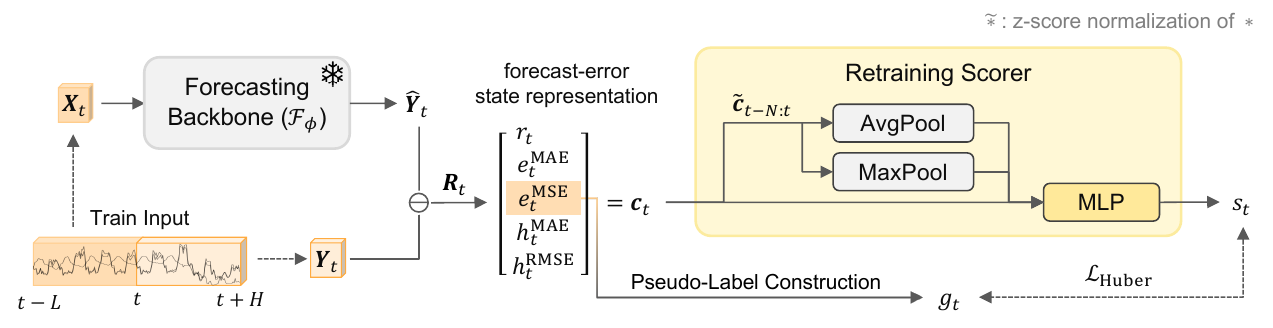}
    \caption{Training Phase}
    \label{fig:training}
  \end{subfigure}
  \begin{subfigure}[b]{0.85\textwidth}
    \vspace{+13pt}
    \centering
    \includegraphics[width=\textwidth]{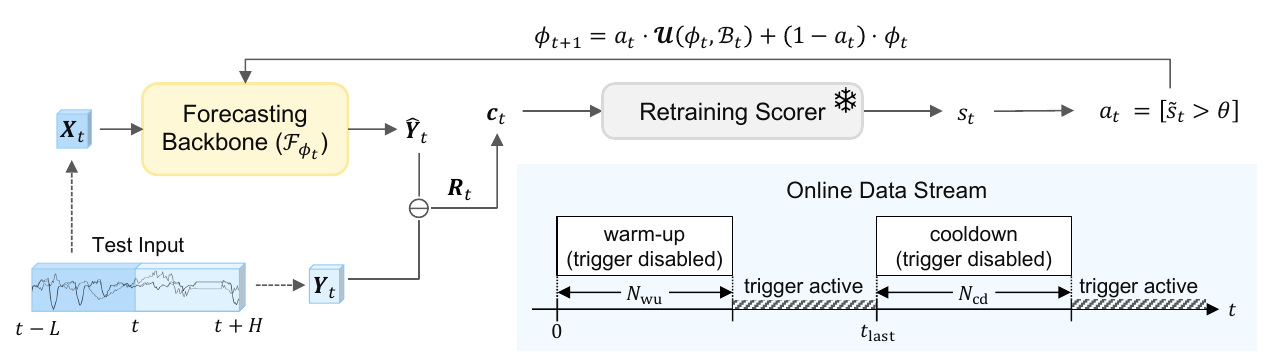}
    \caption{Inference Phase}
    \label{fig:inference}
  \end{subfigure}
  \Description{A two-panel schematic of the proposed retraining method. In both panels, modules shown in gray with a snowflake icon are frozen, while modules shown in light yellow have their parameters updated in the corresponding phase: the scorer is trained in panel (a), and the backbone is retrained in panel (b). The top panel illustrates the offline training phase: a training input window spanning from time $t-L$ to $t+H$ is fed into the frozen forecasting backbone $\gF_{\phi}$, which produces a forecast $\hat{\mY}_t$. The residual $\mR_t = \mY_t -\hat{\mY}_t$ between the ground-truth future segment $\mY_t$ and the forecast is summarized into a five-dimensional forecast-error state vector $\vc_t$ consisting of the average signed residual, the mean absolute error, the mean squared error, the rolling mean absolute error, and the rolling root mean squared error. The mean squared error component of $\vc_t$ is passed through a pseudo-label construction branch that produces a pseudo-label $g_t$. The standardized error state sequence $\mC_t$ over the recent $N$ steps is passed to the trainable retraining scorer, which aggregates the sequence via average pooling and max pooling and concatenates the result with the current-step vector before feeding it into a multi-layer perceptron (MLP) that outputs a scalar retraining score $s_t$, supervised by $g_t$ via the Huber loss. 
  The bottom panel illustrates the online inference phase: the forecasting backbone $\gF_{\phi}$ processes a test input window and produces a forecast whose residual $\mR_t$ forms the current error-state $\vc_t$, which is passed to the frozen retraining scorer to yield a scalar score $s_t$. The score is then online-calibrated and compared against a threshold $\theta$ to produce a binary retraining action $a_t = [\tilde{s}_t > \theta]$. A backbone update equation at the top indicates that when $a_t = 1$, the backbone parameters are updated on a recent training buffer $\mathcal{B}_t$ via the retraining operator $\mathcal{U}$. A timeline at the bottom illustrates the triggering schedule: an initial warm-up period of length $N_{\text{wu}}$, during which triggering is disabled, is followed by an active triggering region, then a cooldown period of length $N_{\text{cd}}$ after the most recent retraining time $t_{\text{last}}$, during which triggering is again disabled, and finally another active triggering region.}
  
  \caption{
  Overall framework of PILOT (Pseudo-label-Informed Learned Online Trigger).
  Gray snowflake modules are frozen while light-yellow modules are updatable in the corresponding phase.
  }
  \label{fig:method}
\end{figure*}


\section{Related Work}
\label{sec:related-work}

\paragraph{\textbf{Selective retraining policies.}}
Despite its practical importance, relatively few studies have directly addressed when models should be retrained.
Zliobaite et al.~\cite{zliobaite2015costsensitive} formulated adaptation as a trade-off between update cost and expected performance gain.
CARA~\cite{mahadevan2024cara} triggers retraining when model staleness cost exceeds retraining cost, whereas UPF~\cite{regol2025upf} predicts future model performance and retrains when predicted quality deteriorates.
These methods infer retraining need from cost, staleness, or predicted performance; 
PILOT instead learns from realized forecast-error dynamics.

\paragraph{\textbf{Drift and forecast-error monitoring.}}
Drift detectors can also serve as update triggers: ADWIN~\cite{bifet2007adwin} monitors changes with adaptive windows, while KSWIN~\cite{raab2020kswin} applies a Kolmogorov--Smirnov test to recent stream windows.
However, detected distribution change need not imply that retraining is beneficial~\cite{regol2025upf}.
Separately, forecast errors have long served as monitoring signals~\cite{trigg1964monitoring}, with recent work detecting model inadequacy directly from forecast-error changes~\cite{grundy2026online} and studying large forecast errors under structural change~\cite{castle2025largeerrors}.
Beyond monitoring, D\textsuperscript{3}A~\cite{zhang2024d3a} used increases in forecasting loss to trigger additional adaptation. 
PILOT learns a retraining trigger from recent error dynamics using future performance degradation as pseudo-label supervision.


\section{Proposed Method}
\label{sec:method}

Figure~\ref{fig:method} summarizes the offline training and online inference phases of PILOT.
During offline scorer training, the pretrained forecasting backbone is frozen, and its completed forecast errors are used to construct error states and a pseudo-label for training a lightweight scorer.
At deployment, the scorer is frozen and uses only completed forecast errors to decide when the backbone should be retrained.

\subsection{Problem Formulation}
\label{sec:method-problem}

We study online multivariate time series forecasting under non-stationary data streams.
Let $\gF_{\phi_t}$ denote the forecasting backbone at decision step $t$, with parameters $\phi_t$.
Given a lookback window $\mX_t \in \sR^{L \times D}$, it produces an $H$-step forecast
\begin{equation}
    \hat{\mY}_t = \gF_{\phi_t}(\mX_t) \in \sR^{H \times D},
\end{equation}
where $L$, $H$, and $D$ denote the lookback length, prediction horizon, and number of variates, respectively, and $\mY_t \in \sR^{H \times D}$ denotes the corresponding future segment.

Once $\mY_t$ is fully observed, the system chooses a binary retraining action $a_t \in \{0,1\}$ using only forecast errors whose full horizons are available.
Let $\mathcal{B}_t$ denote a buffer of recent observations and $\mathcal{U}$ the retraining operator.
The backbone parameters evolve as
\begin{equation}
    \phi_{t+1}
    = a_t\,\mathcal{U}(\phi_t,\mathcal{B}_t)
    +(1-a_t)\phi_t,
\end{equation}
where $a_t=1$ triggers retraining and $a_t=0$ keeps the current backbone unchanged.

Let $\mathcal{E}_t$ denote the history of completed forecast errors observed up to time $t$.
We base the retraining decision on $\mathcal{E}_t$ and seek to learn a decision function $\pi$ that maps this error history to a retraining action,
\begin{equation}
    a_t = \pi(\mathcal{E}_t).
\end{equation}
PILOT realizes $\pi$ by encoding $\mathcal{E}_t$ into a compact forecast-error state representation and learning its relation to future degradation.

\subsection{Forecast-Error State Representation}
\label{sec:method-state-const}

For each completed forecast, we compute the residual matrix
\begin{equation}
    \mR_t = \mY_t - \hat{\mY}_t \in \sR^{H \times D},
\end{equation}
and summarize it over the prediction horizon and variates as
\begin{equation}
\label{eq:error-summaries}
    r_t = \operatorname{Avg}(\mR_t),\quad
    e_t^{\mathrm{MAE}} = \operatorname{Avg}(|\mR_t|),\quad
    e_t^{\mathrm{MSE}} = \operatorname{Avg}(\mR_t^2),
\end{equation}
where $\operatorname{Avg}(\cdot)$ denotes the mean over all entries in $\sR^{H \times D}$.
Here, $r_t$ captures the average signed residual, while
$e_t^{\mathrm{MAE}}$ and $e_t^{\mathrm{MSE}}$ measure the mean absolute error (MAE) and mean squared error (MSE), respectively.

To provide a recent baseline, we additionally compute
\begin{equation}
\label{eq:history-summaries}
    h_t^{\mathrm{MAE}}
    = \frac{1}{K}\sum_{i=t-K+1}^{t} e_i^{\mathrm{MAE}},
    \qquad
    h_t^{\mathrm{RMSE}}
    = \sqrt{\frac{1}{K}\sum_{i=t-K+1}^{t} e_i^{\mathrm{MSE}}},
\end{equation}
and form the per-step state
\begin{equation}
    \vc_t =
    \big[
    r_t,e_t^{\mathrm{MAE}},e_t^{\mathrm{MSE}},
    h_t^{\mathrm{MAE}},h_t^{\mathrm{RMSE}}
    \big]^\top
    \in \sR^5.
\end{equation}
The historical terms $h_t^{\text{MAE}}$ and $h_t^{\text{RMSE}}$ let the scorer interpret current errors relative to a recent baseline rather than in isolation.

Since these feature channels can have different scales across datasets and forecasting backbones, we standardize each channel using statistics computed from the scorer-training split. Let $\boldsymbol{\mu}_{\vc} \in \sR^{5}$ and $\boldsymbol{\sigma}_{\vc} \in \sR^{5}$ denote the feature-wise mean and standard deviation, respectively. The standardized per-step feature vector is defined as
\begin{equation}
    \tilde{\vc}_t = \frac{\vc_t - \boldsymbol{\mu}_{\vc}}{\boldsymbol{\sigma}_{\vc} + \epsilon},
\end{equation}
where $\epsilon$ is a small constant for numerical stability.

Finally, we stack the most recent $N$ 
standardized states:
\begin{equation}
    \mC_t = \big[\tilde{\vc}_{t-N+1};\, \tilde{\vc}_{t-N+2};\, \dots\, ;\, \tilde{\vc}_t\big] \in \sR^{N \times 5}.
\end{equation}
This tensor serves as the input to the retraining scorer described in Section~\ref{sec:scorer}. 
By incorporating the most recent $N$ states, it provides richer temporal context than a single-step snapshot.

\subsection{Pseudo-Label Construction}
\label{sec:method-pseudo-label-const}

Since ground-truth labels indicating retraining necessity are unavailable in practice, we construct an offline pseudo-label $g_t$ on the scorer-training split $\mathcal{D}_{\mathrm{scorer}}$.
We use the change in future mean MSE relative to a recent baseline as the supervision signal, which can be constructed directly from the frozen backbone's error trajectory without counterfactual retraining simulation.

Specifically, mean degradation measures the change in future average MSE relative to a recent baseline, with positive values corresponding to degradation.
This perspective is related to model staleness, which measures the performance cost of keeping an outdated model~\cite{mahadevan2024cara}.

Let $\bar e^{\mathrm{MSE}}_{[a,b)}$ denote the average MSE over the time interval $[a,b)$:
\begin{equation}
    \bar e^{\mathrm{MSE}}_{[a,b)}
    = \frac{1}{b-a}
       \sum_{i=a}^{b-1} e_i^{\mathrm{MSE}}.
\end{equation}
For a current window of length $W_c$ and a future window of length $W_f$, we define
\begin{equation}
    g_t
    = \bar e^{\mathrm{MSE}}_{[t,t+W_f)}
     - \bar e^{\mathrm{MSE}}_{[t-W_c,t)}.
\label{eq:mean-deg}
\end{equation}

Thus, $g_t>0$ indicates that the frozen backbone's future average error exceeds its recent baseline, providing supervision for learning impending degradation.
Following the principle of winsorization~\cite{hastings1947low}, we clip $g_t$ to $[-0.5,2.0]$ before scorer training, affecting fewer than $1\%$ of targets.
We further examine the extensibility of PILOT to alternative pseudo-labels in Section~\ref{sec:ablation} and Appendix~\ref{app:alternative-pseudolabels}.

\subsection{Lightweight Retraining Scorer}
\label{sec:scorer}

Given $\mC_t$ and $g_t$, we train a compact scorer that outputs a retraining score $s_t \in \sR$ at each time step. 
We first aggregate the recent error history by feature-wise average pooling, max pooling, and the latest state:
\begin{equation}
    \vp_t = \big[\operatorname{AvgPool}(\mC_t),\, \operatorname{MaxPool}(\mC_t),\, \tilde{\vc}_t\big] \in \sR^{15}.
\end{equation}
Each component captures a different aspect of the recent error history: average pooling captures the typical recent behavior, max pooling the peak severity, and $\tilde{\vc}_t$ the current state.

We then map $\vp_t$ to the scalar score $s_t$ using a single-hidden-layer multilayer perceptron (MLP) with ReLU activation and dropout:
\begin{equation}
    s_t = \operatorname{MLP}(\vp_t).
\end{equation}

We train the scorer with the Huber loss~\cite{huber1964robust},
\begin{equation}
    \mathcal{L}_{\mathrm{Huber}}
    =
    \frac{1}{|\mathcal{D}_{\mathrm{scorer}}|}
    \sum_{t\in\mathcal{D}_{\mathrm{scorer}}}
    \ell_{\delta}^{\mathrm{Huber}}(s_t,g_t),
\end{equation}
which is robust to occasional large pseudo-label deviations.
To mitigate the underrepresentation of positive pseudo-labels, we oversample scorer-training samples with $g_t>0$~\cite{he2009learning}.
The trained scorer therefore predicts future degradation from error states available at decision time.

\subsection{Online Triggering}
\label{sec:calibration}

At test time, the frozen scorer outputs a raw score $s_t$ from each state tensor $\mC_t$.
We maintain its running mean $\mu_{s,t}$ and standard deviation $\sigma_{s,t}$ using Welford's online update rule~\cite{welford1962note}.
Triggering is disabled until $N_{\mathrm{wu}}$ score observations have been accumulated; afterward, we calibrate
\begin{equation}
\label{eq:calibration}
    \tilde{s}_t
    =
    \frac{s_t-\mu_{s,t}}
         {\sigma_{s,t}+\epsilon}.
\end{equation}
This standardization follows common practice in online monitoring of data streams~\cite{gama2014survey}.

We trigger retraining when the calibrated score exceeds $\theta$, subject to a cooldown of $N_{\mathrm{cd}}$ steps after the previous retraining time $t_{\mathrm{last}}$. 
The final retraining action is therefore defined as
\begin{equation}
\label{eq:trigger}
    a_t
    =
    [n_t \ge N_{\mathrm{wu}}]
    \cdot [\tilde{s}_t > \theta]
    \cdot [t-t_{\mathrm{last}} \ge N_{\mathrm{cd}}],
\end{equation}
where $[\cdot]$ denotes the indicator function and $n_t$ is the number of accumulated scores.
When $a_t=1$, the backbone is warm-start retrained on $\mathcal{B}_t$ according to Section~\ref{sec:method-problem}.
Running score statistics continue to update across retraining events; pseudo-labels are never used at deployment.


\section{Experiments}
\subsection{Experimental Setup}
\paragraph{\textbf{Datasets.}}
In contrast to prior retraining studies that evaluated synthetic or custom-designed streams~\cite{mahadevan2024cara, regol2025upf}, we used eight widely used multivariate forecasting benchmarks, providing a realistic testbed for retraining decisions on naturally evolving real-world time series:
ETTh1 and ETTh2 (hourly, 7 variates), ETTm1 and ETTm2 (15-minute, 7 variates), Electricity (hourly, 321 variates), Exchange (daily, 8 variates), Traffic (hourly, 862 variates), and Weather (10-minute, 21 variates)~\cite{electricity20112014, lai2018lstnet, zhou2021informer, wu2021autoformer}. 

Following the standard split convention adopted in prior long-term forecasting work~\cite{zhou2021informer, wu2021autoformer, wu2023timesnet, liu2024itransformer}, the ETT datasets were split chronologically into training, validation, and test sets at a 60:20:20 ratio, while the other datasets were split at a 70:10:20 ratio. 
Since our method requires an additional split for training the retraining scorer, we further partitioned the standard validation set into two equal halves: one was used as the scorer-training split for constructing error states and training the retraining scorer, and the other was retained as a validation split. 
This yielded four chronologically ordered splits---backbone training, scorer training, validation, and test---with ratios of 60:10:10:20 for ETT datasets and 70:5:5:20 for the others. 
The backbone was pretrained only on the first split and shared throughout all methods, so differences in performance were attributable solely to each method's retraining policy.

\paragraph{\textbf{Backbones.}}
We evaluated PILOT with three forecasting backbones that span different architectural families: DLinear~\cite{zeng2023dlinear} from the linear family; iTransformer~\cite{liu2024itransformer} from the Transformer family; and TimesNet~\cite{wu2023timesnet} from the convolutional family.

\paragraph{\textbf{Baselines.}} 
We compared against six baselines spanning no retraining, scheduled retraining, drift-based triggers, and cost- or prediction-based policies. 
Method-specific hyperparameters were selected on the standard validation split from the candidate sets described below, and the selected configuration was evaluated on the test split.

\begin{itemize}[leftmargin=15pt, topsep=1pt, itemsep=1pt]
  \item \textit{No Retrain} never updates the backbone after initial training.
  
  \item \textit{Periodic} triggers retraining at a fixed interval, with the interval tuned from $\{50, 100, 200\}$ steps.
  
  \item \textit{ADWIN}~\cite{bifet2007adwin} is an adaptive windowing drift detector that automatically adjusts its window size based on observed changes, applied to the per-step MSE summary $e^{\text{MSE}}_t$ defined in Section~\ref{sec:method-state-const}. We tuned the significance level $\delta \in \{0.002, 0.01, 0.05\}$.
  
  \item \textit{KSWIN}~\cite{raab2020kswin} performs a Kolmogorov–Smirnov test over a sliding window of recent $e^{\text{MSE}}_t$ values to detect distributional change. We tuned the significance level $\alpha \in \{0.001, 0.01, 0.05\}$.
  
  \item \textit{CARA}~\cite{mahadevan2024cara} is a cost-aware retraining policy that triggers an update when estimated model staleness exceeds the retraining cost. We tuned both staleness variants ($\mathrm{T}$ and $\mathrm{C}$), with the threshold fixed at $\tau = 0.05$.
  
  \item \textit{UPF}~\cite{regol2025upf} formulates retraining as a future-performance prediction problem using an ElasticNet-based forecaster. We adapted UPF to forecasting by replacing the original predicted quality signal with future MSE under the Gaussian-likelihood setting described in the original paper's Appendix. We tuned the retraining cost penalty $\alpha \in \{0.05, 0.5, 1.0\}$.
\end{itemize}

\paragraph{\textbf{Evaluation metrics.}}
We report test MSE, the number of retraining events (\#RT), average rank, and PILOT's wins/losses against each baseline. 
Lower MSE and rank are better, while lower \#RT indicates fewer model updates; 
MSE ties in win/loss counts are broken by fewer retraining events. 
For each backbone, we also report one-sided Wilcoxon signed-rank tests~\citep{wilcoxon1945test} on 24 paired MSE observations (eight datasets × three shared seeds), pairing PILOT with each baseline to assess whether PILOT yields lower MSE.

\paragraph{\textbf{Implementation details.}}
Across all methods, we used a lookback length of $L=96$ and a prediction horizon of $H=96$.
The cooldown period was set to $N_{\text{cd}}=100$ hours and the retraining buffer length was 1000 steps; 
the same warm-start retraining protocol was used across all methods.
Hour-based hyperparameters were converted to dataset-specific step counts using each sampling interval, preserving a consistent temporal scope across datasets.
Unless otherwise specified, all results were aggregated over three random seeds.

For PILOT, all hyperparameters except the trigger threshold $\theta$ were fixed across datasets and backbones. 
Although $\theta$ remained available for validation, it was fixed to $1.0$ for all results reported in this paper to avoid dataset- or backbone-specific tuning.
The error-state vector used a rolling history window of $K=20$ steps, and the state tensor stacked the most recent $N=24$ standardized vectors. The degradation labels used current and future windows of $W_c = W_f = 48$ hours.
Triggering was disabled during an initial warm-up period of $N_{\mathrm{wu}}=50$ steps, during which online score statistics were accumulated.
The scorer was a single-hidden-layer MLP with a hidden dimension of $64$ and a dropout rate of $0.1$.
The source code is available at \url{https://github.com/Yeryeong9698/PILOT}.

\subsection{Forecasting Performance}
\label{sec:forecast-performance}

\begin{table*}[!t]
\centering
\caption{Forecasting performance across eight datasets and three forecasting backbones. Each cell reports MSE with its standard deviation, followed by the average number of retraining events (\#RT) in parentheses; No Retrain reports MSE only since its \#RT is always zero. \textbf{Bold}: best, \underline{underline}: second-best.}
\label{tab:main}
\setlength{\tabcolsep}{2pt}
\resizebox{\textwidth}{!}{
\begin{tabular}{clccccccc}
\toprule
\textbf{Backbone} & \multicolumn{1}{c}{\textbf{Dataset}} & \textbf{No Retrain} & \textbf{Periodic} & \textbf{ADWIN~(\citeyear{bifet2007adwin})} & \textbf{KSWIN~(\citeyear{raab2020kswin})} & \textbf{CARA~(\citeyear{mahadevan2024cara})} & \textbf{UPF~(\citeyear{regol2025upf})} & \textbf{PILOT~(ours)} \\
\midrule
\multirow{11}{*}{\textbf{DLinear}} & ETTh1 & 0.4572\std{0.0005} & 0.4514\std{0.0003}\,(16.0) & 0.4533\std{0.0002}\,(10.7) & {\ul 0.4506\std{0.0001}\,(32.0)} & 0.4562\std{0.0013}\,(10.3) & 0.4536\std{0.0007}\,(15.3) & \textbf{0.4501\std{0.0003}\,(17.3)} \\
 & ETTh2 & 0.2343\std{0.0085} & \textbf{0.2184\std{0.0007}\,(16.0)} & {\ul 0.2185\std{0.0013}\,(6.7)} & 0.2188\std{0.0009}\,(31.0) & 0.2343\std{0.0085}\,(0.0) & 0.2193\std{0.0008}\,(24.7) & 0.2185\std{0.0009}\,(14.0) \\
 & ETTm1 & 0.3651\std{0.0027} & 0.3606\std{0.0002}\,(35.0) & \textbf{0.3586\std{0.0003}\,(23.3)} & 0.3618\std{0.0009}\,(34.0) & {\ul 0.3594\std{0.0006}\,(23.7)} & 0.3624\std{0.0003}\,(24.0) & 0.3606\std{0.0007}\,(20.7) \\
 & ETTm2 & 0.1756\std{0.0106} & {\ul 0.1603\std{0.0011}\,(35.0)} & 0.1625\std{0.0006}\,(17.3) & 0.1607\std{0.0009}\,(34.0) & 0.1629\std{0.0011}\,(15.3) & 0.1610\std{0.0039}\,(13.3) & \textbf{0.1598\std{0.0031}\,(22.3)} \\
 & ECL & 0.1963\std{0.0000} & \textbf{0.1957\std{0.0000}\,(25.0)} & 0.1963\std{0.0000}\,(6.0) & 0.1963\std{0.0000}\,(50.0) & 0.1962\std{0.0001}\,(9.0) & {\ul 0.1959\std{0.0000}\,(3.0)} & 0.1962\std{0.0002}\,(31.7) \\
 & Exchange & 0.0888\std{0.0000} & 0.0785\std{0.0003}\,(7.0) & 0.0844\std{0.0000}\,(1.0) & 0.0785\std{0.0004}\,(13.0) & {\ul 0.0782\std{0.0001}\,(4.0)} & 0.0888\std{0.0000}\,(0.0) & \textbf{0.0776\std{0.0006}\,(23.0)} \\
 & Weather & 0.1911\std{0.0000} & {\ul 0.1827\std{0.0001}\,(18.0)} & 0.1849\std{0.0001}\,(12.0) & 0.1828\std{0.0002}\,(17.7) & 0.2012\std{0.0009}\,(2.0) & 0.1831\std{0.0001}\,(17.0) & \textbf{0.1819\std{0.0005}\,(14.0)} \\
 & Traffic & 0.6546\std{0.0009} & 0.6416\std{0.0002}\,(17.0) & 0.6424\std{0.0009}\,(7.7) & 0.6420\std{0.0002}\,(32.0) & 0.6425\std{0.0002}\,(20.7) & {\ul 0.6413\std{0.0003}\,(19.0)} & \textbf{0.6410\std{0.0005}\,(15.7)} \\
\cmidrule(){2-9}
 & avg. rank & 6.33 & {\ul 2.73} & 4.13 & 3.81 & 4.85 & 4.15 & \textbf{2.00} \\
 & \# of wins/losses & 8/0 & 6/2 & 6/2 & 8/0 & 6/2 & 7/1 & -- \\
 & Wilcoxon test ($p$) & $<0.001$ & 0.008 & 0.002 & $<0.001$ & $<0.001$ & $<0.001$ & -- \\
\midrule
\multirow{11}{*}{\textbf{iTransformer}} & ETTh1 & 0.4672\std{0.0000} & 0.4616\std{0.0121}\,(16.0) & {\ul 0.4565\std{0.0058}\,(9.3)} & 0.4803\std{0.0222}\,(30.3) & 0.4733\std{0.0159}\,(12.0) & 0.4656\std{0.0091}\,(15.3) & \textbf{0.4459\std{0.0008}\,(19.7)} \\
 & ETTh2 & 0.2468\std{0.0014} & 0.2449\std{0.0015}\,(16.0) & {\ul 0.2441\std{0.0026}\,(8.7)} & 0.2511\std{0.0010}\,(30.7) & 0.2468\std{0.0014}\,(0.0) & 0.2484\std{0.0015}\,(26.0) & \textbf{0.2415\std{0.0017}\,(11.3)} \\
 & ETTm1 & 0.4852\std{0.0260} & {\ul 0.4007\std{0.0057}\,(35.0)} & 0.4023\std{0.0029}\,(25.7) & 0.4086\std{0.0089}\,(33.3) & 0.4033\std{0.0024}\,(27.0) & 0.4086\std{0.0036}\,(25.0) & \textbf{0.3786\std{0.0034}\,(17.0)} \\
 & ETTm2 & 0.1750\std{0.0027} & {\ul 0.1693\std{0.0022}\,(35.0)} & 0.1715\std{0.0015}\,(18.0) & 0.1729\std{0.0035}\,(33.3) & 0.1823\std{0.0062}\,(5.0) & 0.1757\std{0.0038}\,(16.0) & \textbf{0.1680\std{0.0064}\,(15.3)} \\
 & ECL & 0.1661\std{0.0027} & {\ul 0.1646\std{0.0003}\,(25.0)} & 0.1669\std{0.0012}\,(6.3) & 0.1656\std{0.0003}\,(49.7) & 0.1664\std{0.0008}\,(3.7) & 0.1658\std{0.0004}\,(5.3) & \textbf{0.1630\std{0.0006}\,(32.3)} \\
 & Exchange & {\ul 0.0939\std{0.0000}} & 0.1000\std{0.0011}\,(7.0) & \textbf{0.0923\std{0.0005}\,(2.0)} & 0.1085\std{0.0017}\,(13.0) & 0.1039\std{0.0014}\,(8.3) & {\ul 0.0939\std{0.0000}\,(1.0)} & 0.1125\std{0.0123}\,(20.7) \\
 & Weather & 0.1811\std{0.0000} & 0.1803\std{0.0001}\,(18.0) & 0.1800\std{0.0006}\,(14.0) & 0.1816\std{0.0006}\,(17.3) & \textbf{0.1770\std{0.0004}\,(2.0)} & 0.1792\std{0.0005}\,(17.0) & {\ul 0.1788\std{0.0012}\,(13.3)} \\
 & Traffic & {\ul 0.4754\std{0.0000}} & 0.4766\std{0.0002}\,(17.0) & 0.4762\std{0.0001}\,(9.0) & \textbf{0.4753\std{0.0004}\,(32.0)} & 0.4773\std{0.0001}\,(12.0) & 0.4816\std{0.0003}\,(7.0) & 0.4783\std{0.0036}\,(13.7) \\
\cmidrule(){2-9}
 & avg. rank & 4.35 & 3.29 & {\ul 3.21} & 5.13 & 4.73 & 4.92 & \textbf{2.38} \\
 & \# of wins/losses & 6/2 & 6/2 & 6/2 & 6/2 & 5/3 & 7/1 & -- \\
 & Wilcoxon test ($p$) & 0.008 & 0.017 & 0.047 & 0.001 & 0.003 & 0.004 & -- \\
\midrule
\multirow{11}{*}{\textbf{TimesNet}} & ETTh1 & 0.6762\std{0.0041} & 0.6287\std{0.0098}\,(16.0) & {\ul 0.6269\std{0.0035}\,(12.3)} & 0.6619\std{0.0164}\,(32.0) & 0.6396\std{0.0046}\,(10.7) & \textbf{0.6194\std{0.0043}\,(15.0)} & 0.6406\std{0.0091}\,(18.0) \\
 & ETTh2 & \textbf{0.3128\std{0.0135}} & 0.3327\std{0.0093}\,(16.0) & {\ul 0.3170\std{0.0204}\,(8.7)} & 0.3452\std{0.0082}\,(32.0) & \textbf{0.3128\std{0.0135}\,(0.0)} & 0.3432\std{0.0104}\,(30.3) & 0.3270\std{0.0174}\,(15.7) \\
 & ETTm1 & 0.5381\std{0.0000} & 0.3980\std{0.0004}\,(35.0) & 0.4010\std{0.0046}\,(25.7) & 0.3991\std{0.0010}\,(34.0) & 0.4194\std{0.0058}\,(25.3) & {\ul 0.3977\std{0.0013}\,(30.0)} & \textbf{0.3945\std{0.0023}\,(20.0)} \\
 & ETTm2 & 0.1977\std{0.0118} & \textbf{0.1808\std{0.0025}\,(35.0)} & 0.1840\std{0.0042}\,(20.0) & {\ul 0.1810\std{0.0038}\,(34.0)} & 0.1927\std{0.0074}\,(5.0) & 0.1850\std{0.0013}\,(16.7) & 0.1817\std{0.0015}\,(17.7) \\
 & ECL & 0.1731\std{0.0000} & {\ul 0.1619\std{0.0002}\,(25.0)} & 0.1642\std{0.0009}\,(5.3) & 0.1626\std{0.0004}\,(49.7) & 0.1688\std{0.0001}\,(4.0) & 0.1652\std{0.0010}\,(4.7) & \textbf{0.1618\std{0.0003}\,(24.0)} \\
 & Exchange & 0.1828\std{0.0001} & 0.1680\std{0.0002}\,(7.0) & 0.1702\std{0.0023}\,(4.3) & \textbf{0.1656\std{0.0002}\,(13.0)} & {\ul 0.1659\std{0.0001}\,(5.0)} & 0.1828\std{0.0001}\,(0.0) & 0.1659\std{0.0043}\,(9.0) \\
 & Weather & 0.2011\std{0.0057} & 0.1851\std{0.0010}\,(18.0) & {\ul 0.1831\std{0.0013}\,(12.0)} & 0.1856\std{0.0030}\,(17.3) & 0.2009\std{0.0043}\,(2.0) & 0.1839\std{0.0003}\,(17.0) & \textbf{0.1827\std{0.0051}\,(13.7)} \\
 & Traffic & 0.5996\std{0.0009} & 0.4630\std{0.0008}\,(17.0) & 0.4808\std{0.0054}\,(11.3) & \textbf{0.4577\std{0.0013}\,(32.7)} & 0.4894\std{0.0061}\,(8.0) & {\ul 0.4607\std{0.0010}\,(12.7)} & 0.4714\std{0.0126}\,(17.0) \\
\cmidrule(){2-9}
 & avg. rank & 6.17 & {\ul 3.00} & 3.71 & 3.58 & 4.90 & 3.73 & \textbf{2.92} \\
 & \# of wins/losses & 7/1 & 5/3 & 6/2 & 5/3 & 5/3 & 6/2 & -- \\
 & Wilcoxon test ($p$) & $<0.001$ & 0.366 & 0.345 & 0.060 & 0.006 & 0.094 & -- \\
\bottomrule
\end{tabular}
}
\end{table*}

Table~\ref{tab:main} reports MSE, \#RT, and average rank across eight datasets and three backbones.
Although the best-performing policy varied across the settings, PILOT achieved the lowest average rank on all three backbones, with more wins than losses against every baseline.
The Wilcoxon tests further supported this pattern: PILOT was significantly better ($p<0.05$) in 14 of 18 backbone--baseline comparisons; all four nonsignificant comparisons involved TimesNet.

At the individual-dataset level, PILOT obtained the best MSE on five of eight datasets with both DLinear and iTransformer, but on three with TimesNet.
This less consistent advantage on TimesNet may reflect a greater scope for gains from frequent retraining, given its generally higher No Retrain MSE. PILOT may thus be preferable for backbones with competitive forecasting accuracy.

While Periodic and ADWIN remained competitive on certain backbones, neither ranked best across all three.
The more specialized CARA and UPF did not consistently surpass these simpler policies, while KSWIN even ranked below No Retrain on iTransformer.
Overall, these results indicate that PILOT's forecast-error-based supervision provides a reliable basis for retraining decisions across diverse forecasting architectures.

\begin{figure*}[!t]
  \centering
  \includegraphics[width=\textwidth]{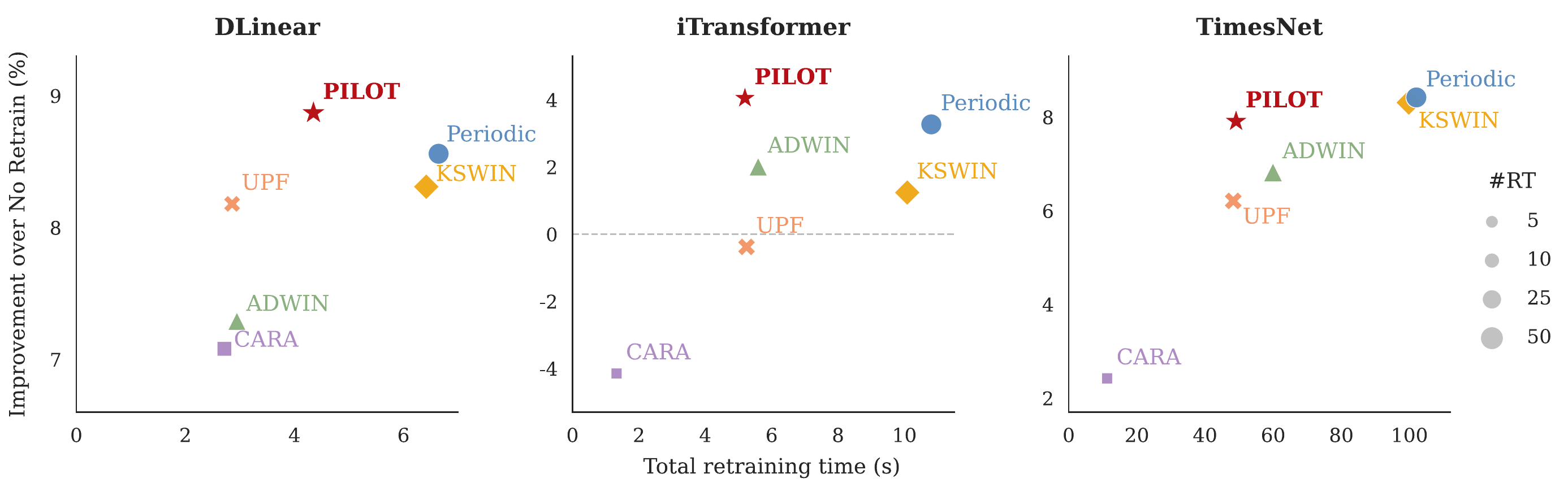}

  \Description{Three scatter plots compare retraining methods on ETTm2 using DLinear, iTransformer, and TimesNet. The x-axis shows total retraining runtime, the y-axis shows relative MSE improvement over No Retrain, and marker size is proportional to the number of retraining events. Methods toward the upper-left achieve greater forecasting improvement with lower retraining runtime.}
  
  \caption{Retraining performance--efficiency trade-off on ETTm2 across the three backbones. 
  The x-axis shows total retraining runtime including decision overhead, the y-axis shows relative MSE improvement over No Retrain (\%), and marker size is proportional to \#RT. 
  Methods closer to the upper-left corner achieve greater forecasting improvement at lower retraining cost. 
  }
  \label{fig:tradeoff}
\end{figure*}

\subsection{Performance--Efficiency Trade-off}
\label{sec:trade-off}

The performance gains of PILOT in Table~\ref{tab:main} were not achieved through indiscriminate retraining. 
Figure~\ref{fig:tradeoff} further examines the performance--efficiency trade-off on ETTm2, one of the longest online test streams among the eight benchmarks and a setting with diverse retraining frequencies across policies.
The figure compares relative improvement over No Retrain against total retraining runtime, with marker size proportional to \#RT; detailed measurements are provided in Appendix~\ref{app:retraining-efficiency}.

PILOT lies on the performance--retraining-time Pareto frontier across all three backbones.
Moreover, total retraining time closely tracked \#RT, while the time per retraining event was broadly similar across methods (Appendix~\ref{app:retraining-efficiency}).
Thus, differences in cumulative retraining cost were driven primarily by how often each policy retrained.
By combining competitive forecasting improvement with moderate retraining cost, PILOT consequently achieved a favorable performance--compute trade-off.

The baselines display two recurring trade-off patterns.
Periodic and KSWIN sit toward the right of each plot, paying high retraining cost for their performance gains.
In contrast, CARA clusters in the lower-left region, keeping retraining cost low but yielding limited improvement;
on iTransformer, CARA even underperformed No Retrain.
The remaining policies occupy intermediate positions but do not reach the upper-left region across all three backbones.
Figure~\ref{fig:tradeoff} therefore confirms PILOT's favorable operational efficiency.

\subsection{Retraining Timing and Robustness}
Beyond aggregate performance and cost, we next isolate the role of retraining timing, examine scorer stability under successive updates, and analyze the resulting trigger behavior.

\begin{table}[t]
\centering
\caption{Effect of retraining timing on DLinear. 
Uniform performs the same number of retraining events as PILOT on each dataset, spacing them uniformly over the test stream.
\textbf{Bold}: best.}
\label{tab:matched_periodic}
\setlength{\tabcolsep}{2pt}
\resizebox{\columnwidth}{!}{
\begin{tabular}{lcccccccc}
\toprule
\textbf{Method}
& \textbf{ETTh1}
& \textbf{ETTh2}
& \textbf{ETTm1}
& \textbf{ETTm2}
& \textbf{ECL}
& \textbf{Exch.}
& \textbf{Weat.}
& \textbf{Traff.} \\
\midrule
No RT
& 0.4572 & 0.2343 & 0.3651 & 0.1756
& 0.1963 & 0.0888 & 0.1911 & 0.6546 \\

Uniform
& 0.5619 & 0.2395 & \textbf{0.3576} & 0.1782
& 0.2041 & 0.0879 & 0.1853 & 0.6416 \\

PILOT
& \textbf{0.4501} & \textbf{0.2185} & 0.3606 & \textbf{0.1598}
& \textbf{0.1962} & \textbf{0.0776} & \textbf{0.1819} & \textbf{0.6410} \\
\bottomrule
\end{tabular}
}
\end{table}

\paragraph{\textbf{Matched-budget timing analysis}}
\label{sec:budget-analysis}

Sections~\ref{sec:forecast-performance} and~\ref{sec:trade-off} show that forecasting accuracy and retraining frequency alone do not reveal whether gains arise from well-timed updates.
To isolate retraining timing from frequency, we compared PILOT with Uniform, which performs the same number of updates as PILOT but distributes them uniformly over the test stream.
We conducted this analysis with DLinear, since Periodic achieved its best average rank among the three backbones in Table~\ref{tab:main}.

As shown in Table~\ref{tab:matched_periodic}, PILOT outperformed Uniform on seven of the eight datasets, except on ETTm1.
Uniform also underperformed No Retrain on four datasets, showing that using the same update budget does not guarantee improvement.
Together, these results provide empirical support for the retraining timing learned from mean-degradation supervision.


\begin{figure}[t]
    \centering
    \includegraphics[width=\columnwidth]{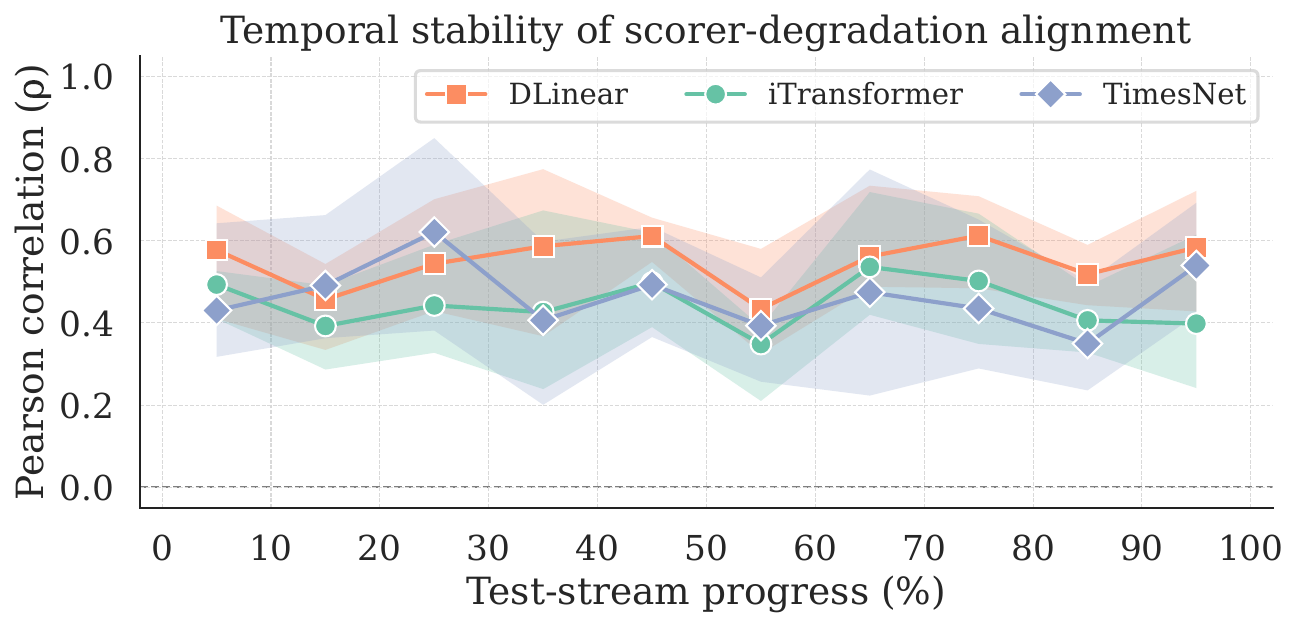}
    \caption{Pearson correlation between $s_t$ and future degradation $g_t^\mathrm{online}$ across the test stream. Lines show means across datasets and seeds; shading denotes the interquartile range.}
    \label{fig:scorer_stability}
\end{figure}

\paragraph{\textbf{Closed-loop robustness}}

The scorer is trained offline with a frozen backbone, whereas online retraining progressively changes the deployed backbone and its error dynamics.
To examine whether the learned signal remains informative under this feedback, we divided each test stream into ten chronological segments and, within each segment, measured the Pearson correlation~\cite{pearson1895note} between $s_t$ and future degradation $g_t^\mathrm{online}$, defined as $g_t$ computed with the active checkpoint $\phi_t$ held fixed over the future window.

Figure~\ref{fig:scorer_stability} shows that the correlation remained consistently positive across all three backbones without systematic decline over the test stream. 
This indicates stable scorer--degradation alignment despite successive backbone updates.


\begin{table}[t]
\centering
\caption{Trigger quality averaged across the eight datasets and three backbones. \textbf{Bold}: best, \underline{underline}: second-best.}
\label{tab:trigger_quality}
\resizebox{\columnwidth}{!}{
\begin{tabular}{lcccc}
\toprule
\multicolumn{1}{c}{\textbf{Method}}
    & \multicolumn{1}{c}{\begin{tabular}[c]{@{}c@{}}\textbf{Trigger}\\\textbf{Precision} \textbf{\footnotesize($\uparrow$)}\end{tabular}}
    & \begin{tabular}[c]{@{}c@{}}\textbf{Trigger}\\\textbf{Need} \textbf{\footnotesize($\uparrow$)}\end{tabular}
    & \begin{tabular}[c]{@{}c@{}}\textbf{Episode}\\\textbf{Hit Rate} \textbf{\footnotesize($\uparrow$)}\end{tabular}
    & \begin{tabular}[c]{@{}c@{}}\textbf{Detection}\\\textbf{Lag Rank} \textbf{\footnotesize($\downarrow$)}\end{tabular} \\
\midrule
Periodic & 0.482 & 0.091 & 0.417 & 3.53 \\
ADWIN    & 0.518 & 0.113 & 0.243 & 3.73 \\
KSWIN    & 0.514 & 0.088 & \textbf{0.574} & \underline{2.66} \\
CARA     & \underline{0.588} & \underline{0.255} & 0.233 & 4.61 \\
UPF      & 0.437 & 0.052 & 0.353 & 4.04 \\
\midrule
PILOT    & \textbf{0.707} & \textbf{0.356} & \underline{0.468} & \textbf{2.43} \\
\bottomrule
\end{tabular}
}
\end{table}

\paragraph{\textbf{Trigger quality analysis}}

The matched-budget results showed that update frequency alone cannot explain forecasting performance.
We therefore evaluated whether each policy's triggers aligned with realized forecast degradation.
Using the No Retrain error stream as a common reference, we defined degradation as an increase in future average error relative to the recent window.

The four metrics capture two complementary perspectives: trigger selectivity and degradation-episode coverage.
\textbf{Trigger precision} measures the fraction of triggers occurring during degradation, while \textbf{trigger need} measures the average normalized degradation magnitude at triggered steps.
\textbf{Episode hit rate} measures the fraction of contiguous degradation episodes containing at least one trigger, whereas \textbf{detection lag rank} measures how quickly a method triggers after the start of a hit episode.
Detection lags are converted to per-dataset ranks before aggregation to account for different sampling rates.%
\footnote{Trigger-conditioned metrics omit settings with no triggers; missed episodes count as zero hits, and methods with no hits receive the worst detection lag rank.}

Table~\ref{tab:trigger_quality} reveals a clear trade-off among the baselines.
CARA is highly selective but misses many degradation episodes and reacts late, whereas KSWIN provides broader and faster coverage with lower selectivity.
UPF shows the weakest selectivity and only moderate episode coverage, while Periodic and ADWIN occupy intermediate positions.
This balanced behavior of Periodic and ADWIN is consistent with their relatively strong forecasting performance in Table~\ref{tab:main}.

PILOT is the only method to rank within the top two on all four metrics: it achieves the best trigger precision, trigger need, and detection lag rank, together with the second-best episode hit rate.
Thus, PILOT combines selective triggering with broad and timely coverage of degradation episodes.

\begin{figure}[t]
  \centering
  \begin{subfigure}[b]{0.495\textwidth}
    \centering
    \includegraphics[width=\textwidth]{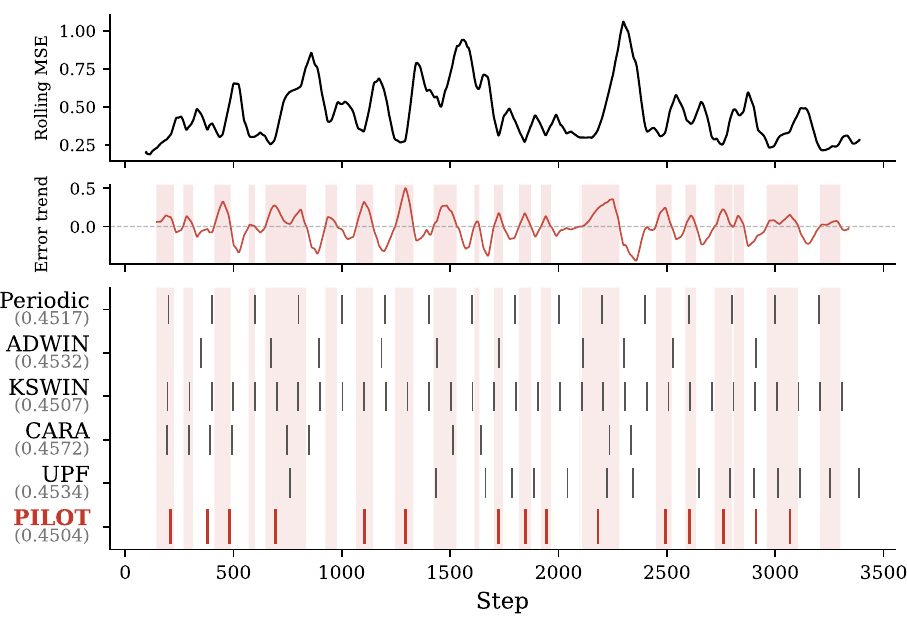}
    \caption{DLinear on ETTh1}
    \label{fig:case_dlinear}
  \end{subfigure}
  \begin{subfigure}[b]{0.495\textwidth}
    \centering
    \includegraphics[width=\textwidth]{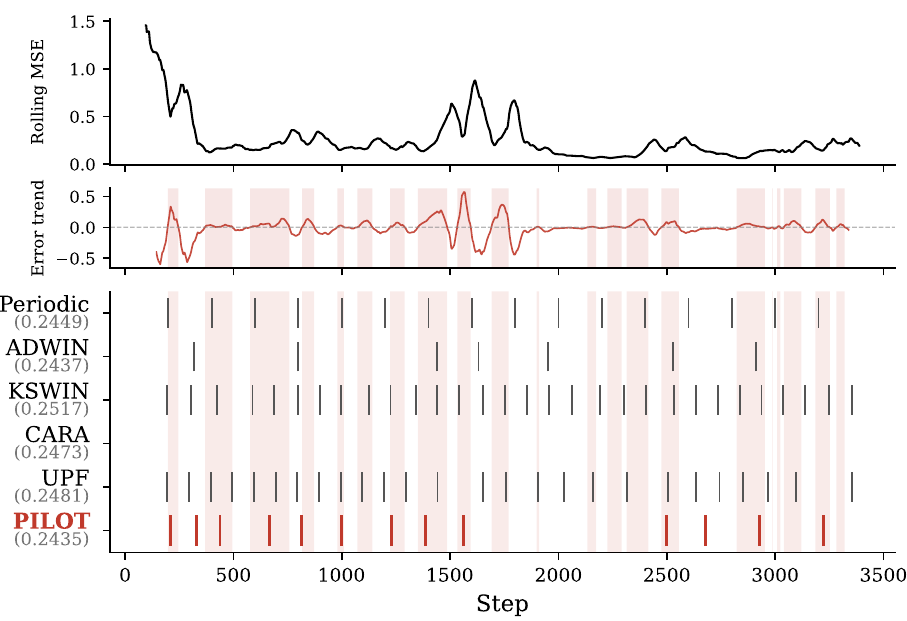}
    \caption{iTransformer on ETTh2}
    \label{fig:case_itrans}
  \end{subfigure}
  \Description{Two three-panel figures illustrate retraining behavior over time. The first set shows ETTh1 with DLinear: the rolling MSE fluctuates with several distinct surges between steps 500 and 2300; the error trend panel shows positive values during these surges, with shaded regions marking intervals of positive forecast degradation; PILOT concentrates triggers within shaded regions, and it achieves the lowest MSE in this case.
  The second set shows ETTh2 with iTransformer: the rolling MSE starts high (~1.4) and gradually decreases, with persistent surges around steps 1400 to 1800; the error trend shows alternating positive and negative regions; PILOT achieves the lowest MSE on this seed while triggering substantially fewer times than others around steps 1600 to 2500.}
  \caption{Trigger visualization of each retraining method for (\subref{fig:case_dlinear}) DLinear on ETTh1 and (\subref{fig:case_itrans}) iTransformer on ETTh2.}
  \label{fig:case_study}
\end{figure}

\paragraph{\textbf{Trigger timing analysis.}}

Figure~\ref{fig:case_study} complements Table~\ref{tab:trigger_quality} with temporal examples for DLinear--ETTh1 and iTransformer--ETTh2.
Within each panel, the top row shows the rolling MSE of the No Retrain baseline; the middle row shows the corresponding degradation trend; and the bottom row shows each method's trigger events for one random seed, with the corresponding test MSE in parentheses.
Shaded regions in the middle and bottom rows mark intervals with positive forecast degradation.
These intervals provide a reference for assessing how each method's triggers align with forecast degradation.

As shown in Figure~\ref{fig:case_dlinear}, the ETTh1 rolling MSE exhibits several distinct surges, including repeated bursts around steps 1300--1800 and a sharp spike near 2300.
Periodic and KSWIN trigger across both degradation and non-degradation intervals, whereas CARA triggers comparatively sparsely.
PILOT concentrates its updates around major degradation intervals and achieves the lowest MSE on this seed.

In Figure~\ref{fig:case_itrans}, the rolling MSE peaks early (around 1.4 within the first 200 steps) and gradually decreases, with persistent surges around steps 1400--1800.
KSWIN and UPF trigger densely over the entire test stream, while CARA never triggers.
PILOT remains comparatively sparse after the main degradation interval and again achieves the lowest MSE.
Together, these cases visually reinforce the selective yet responsive behavior summarized in Table~\ref{tab:trigger_quality}.


\subsection{Ablation and Alternative Pseudo-Labels}
\label{sec:ablation}

\begin{table}[t]
\centering
\setlength{\tabcolsep}{4pt}
\caption{Component ablation for PILOT on TimesNet. Parentheses report relative MSE change from PILOT.}
\label{tab:ablation}
\resizebox{\columnwidth}{!}{
\begin{tabular}{lcccc}
\toprule
\multicolumn{1}{c}{\textbf{Dataset}}
 & \textbf{PILOT}
 & \textbf{w/o Scorer}\textsuperscript{\dag}
 & \textbf{w/o History}
 & \textbf{w/o Stacking} \\
\midrule
ETTh1    & 0.6406 & 0.6893\,($+7.61\%$) & 0.6928\,($+8.14\%$) & 0.7004\,($+9.34\%$) \\
ETTh2    & 0.3270 & 0.3294\,($+0.74\%$) & 0.3320\,($+1.54\%$) & 0.3337\,($+2.04\%$) \\
ETTm1    & 0.3945 & 0.3973\,($+0.71\%$) & 0.4034\,($+2.26\%$) & 0.4024\,($+2.00\%$) \\
ETTm2    & 0.1817 & 0.1891\,($+4.06\%$) & 0.1810\,($-0.38\%$) & 0.1859\,($+2.29\%$) \\
ECL      & 0.1618 & 0.1708\,($+5.55\%$) & 0.1691\,($+4.53\%$) & 0.1697\,($+4.90\%$) \\
Exchange & 0.1659 & 0.1624\,($-2.14\%$) & 0.1678\,($+1.13\%$) & 0.1729\,($+4.21\%$) \\
Weather  & 0.1827 & 0.1866\,($+2.13\%$) & 0.1861\,($+1.86\%$) & 0.1858\,($+1.70\%$) \\
Traffic  & 0.4714 & 0.4796\,($+1.73\%$) & 0.4800\,($+1.83\%$) & 0.4748\,($+0.71\%$) \\
\midrule
\textbf{avg. $\Delta\%$} & \textemdash & $+2.55\%$ & $+2.61\%$ & $+3.40\%$ \\
\textbf{\# of wins} & \textemdash & 7 & 7 & 8 \\
\bottomrule
\end{tabular}
}
\vspace{1pt}
\parbox{0.98\columnwidth}{\footnotesize \textsuperscript{\dag}This future-informed comparison directly thresholds the future-derived retraining label, which is unavailable at deployment and introduces information leakage.}
\end{table}

\paragraph{\textbf{Component ablation.}}

To identify which components are responsible for the gains of PILOT, we conducted an ablation study using the mean-degradation formulation with TimesNet across all eight datasets. 
Table~\ref{tab:ablation} reports MSE and the relative MSE increase from PILOT.
We compared PILOT against three ablated variants: \textbf{w/o Scorer}, which removes the learned scorer and directly thresholds the pseudo-label; \textbf{w/o History}, which removes the rolling historical summaries ($h_t^{\text{MAE}}$, $h_t^{\text{RMSE}}$) from the state; and \textbf{w/o Stacking}, which uses only the current-step state without temporal stacking.

PILOT outperformed its ablated counterparts in 22 of the 24 dataset--variant comparisons.
Removing the learned scorer increased MSE by $2.55\%$ on average, with the largest increase on ETTh1 ($+7.61\%$).
Notably, w/o Scorer is a future-informed method that directly thresholds a pseudo-label unavailable at deployment; PILOT nonetheless outperformed it on seven of eight datasets.
This result is consistent with a denoising or regularization effect of scorer learning and suggests that the learned scorer can yield more reliable retraining decisions than direct thresholding.

Removing historical features increased MSE by $2.61\%$ on average and degraded seven of eight datasets, showing the value of interpreting current errors against a recent baseline.
Removing temporal stacking caused the largest average increase ($+3.40\%$) and degraded all eight datasets, confirming that recent error dynamics are more informative than a single-step state.
Therefore, the three mechanisms are complementary: the scorer converts noisy pseudo-labels into reliable decisions, historical features provide a local reference for error severity, and stacking captures recent error dynamics.

\begin{table}[t]
\centering
\caption{Descriptive average MSE of alternative pseudo-label designs. Formatting indicates their average-rank positions relative to the baselines. \textbf{Bold}: outperforms the best baseline; \underline{underline}: outperforms the second-best.}
\label{tab:pseudolabel-variants}
\setlength{\tabcolsep}{2pt}
\resizebox{\columnwidth}{!}{
\begin{tabular}{lcccc}
\toprule
\textbf{Variant} & \textbf{DLinear} & \textbf{iTransformer} & \textbf{TimesNet}
& {\small \begin{tabular}[c]{@{}c@{}}\textbf{Counterfactual}\\\textbf{RT}\textsuperscript{\dag}\end{tabular}} \\
\midrule
Mean (main) & \textbf{0.2857} & \textbf{0.2708} & \textbf{0.3157} & No \\
Cumulative & \textbf{0.2856} & \textbf{0.2730} & {\ul 0.3164} & No \\
Relative & \textbf{0.2857} & \textbf{0.2714} & {\ul 0.3189} & No \\
Binary util. & {\ul 0.2872} & \textbf{0.2733} & 0.3150 & Yes \\
Ordinal util. & 0.2878 & \textbf{0.2755} & \textbf{0.3137} & Yes \\
Voting & {\ul 0.2873} & {\ul 0.2932} & 0.3191 & Yes \\
\bottomrule
\end{tabular}
}
\parbox{0.98\columnwidth}{
  \vspace{1pt}
  \footnotesize 
  \textsuperscript{\dag}Additional offline retraining simulations are required only to construct utility-based labels; they are not used during online inference. 
  }
\end{table}
\paragraph{\textbf{Alternative pseudo-labels.}}

We additionally evaluated cumulative and relative degradation, binary and ordinal retraining utility, and a voting ensemble of the five scorers. Table~\ref{tab:pseudolabel-variants} summarizes their average MSE and computational requirements; their definitions and detailed analysis are provided in Appendix~\ref{app:alternative-pseudolabels}.

The extension study shows that PILOT is not tied to a single supervision signal.
The degradation-based variants yield similar average MSE on DLinear and iTransformer, while Ordinal Utility performs best on TimesNet.
Overall, these results show that the framework accommodates alternative notions of retraining necessity, while Mean Degradation provides a simple and consistently effective default without counterfactual retraining simulation.

\section{Conclusion}
We proposed PILOT, a pseudo-label-based online retraining method that constructs supervision from realized forecast errors. 
PILOT uses future mean forecast-error degradation as a pseudo-label and trains a lightweight scorer on recent error states to produce retraining decisions. 
It remains backbone-agnostic, requiring no modification to the forecasting model.

Across the eight benchmarks and three backbones, PILOT achieved the lowest average rank among retraining policies while maintaining a favorable performance--retraining-time trade-off. 
Matched-budget and trigger analyses further showed that its gains arise from selective and timely retraining rather than merely from more frequent updates. 
Closed-loop analysis showed that scorer--degradation alignment remained stable despite successive backbone updates, while an ablation study confirmed that scorer learning, historical context, and temporal stacking each play distinct roles in producing selective retraining decisions.

More broadly, these results suggest that forecast errors can serve not only as monitoring signals but also as effective sources of supervision for learning retraining decisions. 
Experiments with alternative pseudo-labels further showed that PILOT is not tied to a single supervision design, while mean degradation provides a simple and consistently effective default without requiring counterfactual retraining simulation.

While PILOT showed promising results, several directions remain open. 
Its pseudo-label relies on fixed-length current and future windows, which may limit responsiveness to abrupt performance degradation. 
PILOT also supports only binary decisions, whereas richer actions such as recalibration, partial fine-tuning, or cost-aware update selection may further improve adaptation under evolving data streams.

\appendix
\section{Retraining Efficiency}
\label{app:retraining-efficiency}

Table~\ref{tab:retraining-efficiency-ettm2} reports \#RT, total retraining time including decision overhead, time per RT, and maximum GPU memory averaged over three seeds, supporting the ETTm2 performance--efficiency comparison in Figure~\ref{fig:tradeoff}.
Within each backbone, peak GPU memory and time per retraining event are broadly comparable across methods.
Accordingly, total retraining cost is driven primarily by how often retraining is invoked, reinforcing the importance of deciding \emph{when} to retrain---the central focus of this work.

\begin{table}[t]
\small
\centering
\caption{Detailed retraining efficiency on ETTm2.}
\label{tab:retraining-efficiency-ettm2}
\setlength{\tabcolsep}{2.5pt}
\resizebox{0.98\columnwidth}{!}
{
\begin{tabular}{clrrrr}
\toprule
\textbf{Backbone}
 & \multicolumn{1}{c}{\textbf{Method}}
 & \multicolumn{1}{c}{\textbf{\#RT}}
& \begin{tabular}[c]{@{}c@{}}\textbf{Total RT}\\\textbf{Time (s)}\end{tabular}
& \begin{tabular}[c]{@{}c@{}}\textbf{Time / \#RT}\\\textbf{(s)}\end{tabular}
& \begin{tabular}[c]{@{}c@{}}\textbf{GPU Mem.}\\\textbf{(MB)}\end{tabular} \\
\midrule
\multirow{6}{*}{\textbf{DLinear}}
 & Periodic & 35.0 & 6.644 & 0.190 & 233.4 \\
 & ADWIN    & 17.3 & 2.944 & 0.170 & 233.5 \\
 & KSWIN    & 34.0 & 6.418 & 0.189 & 233.7 \\
 & CARA     & 15.3 & 2.716 & 0.177 & 233.0 \\
 & UPF      & 13.3 & 2.857 & 0.214 & 233.8 \\
\cmidrule(){2-6}
 & PILOT    & 22.3 & 4.349 & 0.195 & 232.6 \\
\midrule
\multirow{6}{*}{\textbf{iTransformer}}
 & Periodic & 35.0 & 10.810 & 0.309 & 1336.0 \\
 & ADWIN    & 18.0 & 5.595 & 0.311 & 1342.8 \\
 & KSWIN    & 33.3 & 10.085 & 0.303 & 1334.1 \\
 & CARA     & 5.0 & 1.326 & 0.265 & 1336.1 \\
 & UPF      & 16.0 & 5.243 & 0.328 & 1330.6 \\
\cmidrule(){2-6}
 & PILOT    & 15.3 & 5.195 & 0.339 & 1330.6 \\
\midrule
\multirow{6}{*}{\textbf{TimesNet}}
 & Periodic & 35.0 & 102.016 & 2.915 & 39.6 \\
 & ADWIN    & 20.0 & 59.941 & 2.997 & 39.2 \\
 & KSWIN    & 34.0 & 99.807 & 2.935 & 39.9 \\
 & CARA     & 5.0 & 11.272 & 2.254 & 39.7 \\
 & UPF      & 16.7 & 48.280 & 2.897 & 39.2 \\
\cmidrule(){2-6}
 & PILOT    & 17.7 & 49.115 & 2.780 & 39.2 \\
\bottomrule
\end{tabular}
}
\end{table}
\section{Alternative Pseudo-Labels}
\label{app:alternative-pseudolabels}

PILOT constructs retraining supervision from future mean degradation, as described in Section~\ref{sec:method-pseudo-label-const}. 
To examine its extensibility, Section~\ref{sec:ablation} evaluates the same scorer with alternative pseudo-labels designed with degradation- and utility-based signals; their definitions and detailed results are provided below.

\subsection{Alternative Pseudo-Label Definitions}

The alternative single-label variants capture two notions of retraining necessity: (i) future degradation under the unchanged backbone~\cite{mahadevan2024cara, page1954cusum}, and (ii) utility from immediate retraining~\cite{regol2025upf, zliobaite2015costsensitive}.

\subsubsection{Cumulative degradation}
Cumulative degradation focuses on the accumulated future deterioration by summing future MSE deviations from the recent baseline:
\begin{equation}
g_t^{\text{cum}} = \sum_{i=t}^{t+W_f-1} \left( e_i^{\text{MSE}} - \bar{e}^{\text{MSE}}_{[t-W_c,\, t)} \right).
\end{equation}
This aligns with cumulative-sum monitoring statistics~\cite{page1954cusum}.

\subsubsection{Relative degradation}
Relative degradation addresses scale differences by normalizing mean degradation by the recent error baseline:
\begin{equation}
    g_t^{\text{rel}} = \frac{\bar{e}^{\text{MSE}}_{[t,\, t+W_f)} - \bar{e}^{\text{MSE}}_{[t-W_c,\, t)}}{\bar{e}^{\text{MSE}}_{[t-W_c,\, t)} + \epsilon}.
\end{equation}
This normalized formulation is closely related to the relative staleness cost used in cost-aware retraining~\cite{mahadevan2024cara}. 
Both extensions use the same clipping rule as mean degradation.

\subsubsection{Utility-based labels}
Utility-based labels assess immediate retraining benefits by comparing future losses under stale and retrained backbones, following cost-sensitive formulations~\cite{zliobaite2015costsensitive,regol2025upf}.

At each candidate time step $t$, we compare keeping the current backbone unchanged with retraining it on the recent buffer. 
We then define retraining utility as
\begin{equation}
    u_t = \bar{e}^{\text{MSE}}_{\text{stale},\, t} - \bar{e}^{\text{MSE}}_{\text{retrained},\, t},
\end{equation}
where $\bar{e}^{\text{MSE}}_{\text{stale},\, t}$ and $\bar{e}^{\text{MSE}}_{\text{retrained},\, t}$ are the average future errors resulting from the stale case and the retraining case, respectively. 
A positive value indicates that immediate retraining improves future forecasting performance. 
These labels require counterfactual retraining only offline and do not affect online inference.

\subsubsection{Binary retraining utility}
The binary retraining utility label indicates whether immediate retraining yields a positive utility:
\begin{equation}
    g_t^{\text{bin}} = [u_t > 0].
\end{equation}

\subsubsection{Ordinal retraining utility}
An ordinal version separates the retraining benefit into low-, medium-, and high-benefit cases. 
We discretize the scalar utility into three ordinal levels:
\begin{equation}
    g_t^{\text{ord}} =
    \begin{cases}
        0, & u_t \le Q_{50}, \\
        1, & Q_{50} < u_t \le Q_{80}, \\
        2, & u_t > Q_{80},
    \end{cases}
\end{equation}
where $Q_{50}$ and $Q_{80}$ are empirical utility quantiles estimated from the scorer-training split, following a simple quantile-based discretization strategy~\cite{dougherty1995supervised}. 
Unlike continuous degradation labels, these discrete utility labels are used as-is without clipping.

\subsubsection{Voting ensemble}
We additionally combine the five pseudo-label scorers by majority voting on a shared backbone trajectory, following hard-voting strategies in ensemble learning~\cite{dietterich2000ensemble,kuncheva2004combining}. 
At each step $t$, the five frozen scorers run in parallel on the same state tensor $\mC_t$. 
Each scorer $m$ is independently calibrated as in Equation~\ref{eq:calibration} and thresholded to produce a binary vote $a_t^{(m)}$. 
Retraining is triggered when at least three of the five scorers agree:
\begin{equation}
\label{eq:voting}
a_t = \!\left[\sum_{m=1}^{5} a_t^{(m)} \ge 3\right],
\end{equation}
subject to the same cooldown constraint as in Equation~\ref{eq:trigger}.

\begin{table}[t]
\centering
\caption{Dataset-level MSE of the alternative pseudo-label variants, averaged over three seeds. }
\label{tab:pseudolabel-variants-full}
\scriptsize
\setlength{\tabcolsep}{2pt}
\resizebox{\columnwidth}{!}{
\begin{tabular}{clccccc}
\toprule
\textbf{Backbone} & \textbf{Dataset}
& \makecell{\textbf{Cum.}\\\textbf{deg.}}
& \makecell{\textbf{Rel.}\\\textbf{deg.}}
& \makecell{\textbf{Binary}\\\textbf{util.}}
& \makecell{\textbf{Ordinal}\\\textbf{util.}}
& \textbf{Voting} \\
\midrule
\multirow{8}{*}{\textbf{DLinear}}
 & ETTh1    & 0.4503 & 0.4503 & 0.4532 & 0.4537 & 0.4510 \\
 & ETTh2    & 0.2170 & 0.2182 & 0.2189 & 0.2240 & 0.2178 \\
 & ETTm1    & 0.3574 & 0.3574 & 0.3601 & 0.3578 & 0.3601 \\
 & ETTm2    & 0.1605 & 0.1605 & 0.1629 & 0.1620 & 0.1618 \\
 & ECL      & 0.1962 & 0.1962 & 0.1956 & 0.1958 & 0.1960 \\
 & Exchange & 0.0771 & 0.0777 & 0.0790 & 0.0805 & 0.0853 \\
 & Weather  & 0.1850 & 0.1836 & 0.1862 & 0.1860 & 0.1855 \\
 & Traffic  & 0.6412 & 0.6414 & 0.6419 & 0.6427 & 0.6412 \\
\midrule
\multirow{8}{*}{\textbf{iTransformer}}
 & ETTh1    & 0.4475 & 0.4464 & 0.4513 & 0.4508 & 0.4860 \\
 & ETTh2    & 0.2450 & 0.2445 & 0.2465 & 0.2459 & 0.2659 \\
 & ETTm1    & 0.3881 & 0.3791 & 0.3827 & 0.4009 & 0.5007 \\
 & ETTm2    & 0.1662 & 0.1656 & 0.1657 & 0.1658 & 0.1724 \\
 & ECL      & 0.1635 & 0.1634 & 0.1635 & 0.1636 & 0.1701 \\
 & Exchange & 0.1193 & 0.1193 & 0.1193 & 0.1205 & 0.0914 \\
 & Weather  & 0.1793 & 0.1768 & 0.1801 & 0.1785 & 0.1816 \\
 & Traffic  & 0.4753 & 0.4763 & 0.4774 & 0.4778 & 0.4772 \\
\midrule
\multirow{8}{*}{\textbf{TimesNet}}
 & ETTh1    & 0.6349 & 0.6366 & 0.5977 & 0.6221 & 0.6311 \\
 & ETTh2    & 0.3299 & 0.3309 & 0.3245 & 0.3139 & 0.3354 \\
 & ETTm1    & 0.4051 & 0.3987 & 0.4020 & 0.4121 & 0.4043 \\
 & ETTm2    & 0.1807 & 0.1815 & 0.1819 & 0.1797 & 0.1829 \\
 & ECL      & 0.1613 & 0.1619 & 0.1651 & 0.1631 & 0.1612 \\
 & Exchange & 0.1661 & 0.1682 & 0.1721 & 0.1627 & 0.1776 \\
 & Weather  & 0.1839 & 0.1834 & 0.1848 & 0.1829 & 0.1863 \\
 & Traffic  & 0.4687 & 0.4899 & 0.4917 & 0.4733 & 0.4740 \\
\bottomrule
\end{tabular}
}
\end{table}

\subsection{Alternative Pseudo-Label Performance}

The full comparison evaluates Mean Degradation together with four single-label extensions and the voting ensemble. Table~\ref{tab:pseudolabel-variants-full} reports the dataset-level MSE of the five extensions, complementing the across-dataset averages in Table~\ref{tab:pseudolabel-variants}.

Table~\ref{tab:pseudolabel-variants-full} shows substantial dataset-level variation across the extensions. 
Cumulative and Relative Degradation attain the lowest MSE on most DLinear datasets, while Relative Degradation is lowest on six of eight iTransformer datasets. 
TimesNet exhibits a more heterogeneous pattern: Ordinal Utility is lowest on four datasets, with the remaining datasets favoring different variants. 
Voting can be particularly effective in specific settings. For example, on iTransformer--Exchange, it is the only PILOT variant to achieve a lower MSE than No Retrain (0.0914 vs.\ 0.0939).


\begin{acks}
This work was supported by the \grantsponsor{NRF}{National Research Foundation of Korea}{https://doi.org/10.13039/501100003725} (NRF) grant funded by the Korea government (MSIT) (\grantnum{NRF}{RS-2026-25477016}).
\end{acks}

\section*{Ethical Considerations}
All experiments use publicly available time series benchmarks containing no personally identifiable or sensitive individual-level information, and involve no human or animal subjects, protected attributes, or personal data collection. PILOT is a general-purpose retraining framework not designed for sensitive or safety-critical applications, and we identify no foreseeable risks involving privacy, unfair treatment, surveillance, or malicious use. We provide source code and document the experimental settings to support transparency and reproducibility.

The authors used Anthropic’s Claude for code implementation, debugging, figure and table generation, and manuscript refinement, and OpenAI’s ChatGPT for English language refinement. The authors independently determined the research direction, methodology, and experimental procedures, interpreted the results, and verified all AI-assisted outputs and scientific claims, taking full responsibility for the work.



\bibliographystyle{ACM-Reference-Format}
\balance
\bibliography{sample-base}

@String{Computing = "Computing" }

@String{Computer = "{IEEE} Computer" }

@String{Macmillan = "Macmillan" }

@String{Springer = "Springer-Verlag" }

@String{ACM = "Association for Computing Machinery"}

@String{ICML = "International Conference on Machine Learning"}

@String{SIAM = "Society for Industrial and Applied Mathematics"}

@String{PMLR = "Proceedings of Machine Learning Research"}

@misc{zhang2024d3a,
  title         = {Addressing Concept Shift in Online Time Series Forecasting: Detect-then-Adapt},
  author        = {Zhang, YiFan and Chen, Weiqi and Zhu, Zhaoyang and Qin, Dalin and Sun, Liang and Wang, Xue and Wen, Qingsong and Zhang, Zhang and Wang, Liang and Jin, Rong},
  year          = {2024},
  eprint        = {2403.14949},
  archivePrefix = {arXiv},
  primaryClass  = {cs.LG},
  doi           = {10.48550/arXiv.2403.14949}
}

@inproceedings{regol2025upf,
  title     = {When to Retrain a Machine Learning Model},
  author    = {Regol, Florence and Schwinn, Leo and Sprague, Kyle and Coates, Mark and Markovich, Thomas},
  booktitle = {Proceedings of the 42nd International Conference on Machine Learning},
  series    = {Proceedings of Machine Learning Research},
  volume    = {267},
  pages     = {51369--51404},
  publisher = {PMLR},
  year      = {2025}
}

@article{zliobaite2015costsensitive,
  title   = {Towards Cost-Sensitive Adaptation: When Is It Worth Updating Your Predictive Model?},
  author  = {Zliobaite, Indre and Budka, Marcin and Stahl, Frederic},
  journal = {Neurocomputing},
  volume  = {150},
  pages   = {240--249},
  year    = {2015},
  doi     = {10.1016/j.neucom.2014.05.084}
}

@article{mahadevan2024cara,
  title   = {Cost-Aware Retraining for Machine Learning},
  author  = {Mahadevan, Ananth and Mathioudakis, Michael},
  journal = {Knowledge-Based Systems},
  volume  = {293},
  pages   = {111610},
  year    = {2024},
  doi     = {10.1016/j.knosys.2024.111610},
  publisher = {Elsevier}
}

@article{zanotti2025retrainfreq,
  title   = {On the Retraining Frequency of Global Models in Retail Demand Forecasting},
  author  = {Zanotti, Marco},
  journal = {Machine Learning with Applications},
  volume  = {22},
  pages   = {100769},
  year    = {2025},
  doi     = {10.1016/j.mlwa.2025.100769}
}

@inproceedings{bifet2007adwin,
  title     = {Learning from Time-Changing Data with Adaptive Windowing},
  author    = {Bifet, Albert and Gavald{\`a}, Ricard},
  booktitle = {Proceedings of the 2007 SIAM International Conference on Data Mining},
  pages     = {443--448},
  year      = {2007},
  publisher = {Society for Industrial and Applied Mathematics},
  address   = {Philadelphia, PA},
  doi       = {10.1137/1.9781611972771.42}
}

@article{raab2020kswin,
  title   = {Reactive Soft Prototype Computing for Concept Drift Streams},
  author  = {Raab, Christoph and Heusinger, Moritz and Schleif, Frank-Michael},
  journal = {Neurocomputing},
  volume  = {416},
  pages   = {340--351},
  year    = {2020},
  doi     = {10.1016/j.neucom.2019.11.111}
}

@article{trigg1964monitoring,
  title   = {Monitoring a Forecasting System},
  author  = {Trigg, D. W.},
  journal = {Operational Research Quarterly},
  volume  = {15},
  number  = {3},
  pages   = {271--274},
  year    = {1964},
  doi     = {10.1057/jors.1964.48}
}

@article{page1954cusum,
  title   = {Continuous Inspection Schemes},
  author  = {Page, E. S.},
  journal = {Biometrika},
  volume  = {41},
  number  = {1--2},
  pages   = {100--115},
  year    = {1954},
  doi     = {10.1093/biomet/41.1-2.100}
}

@article{gama2014survey,
  title   = {A Survey on Concept Drift Adaptation},
  author  = {Gama, Joao and Zliobaite, Indre and Bifet, Albert and Pechenizkiy, Mykola and Bouchachia, Abdelhamid},
  journal = {ACM Computing Surveys},
  volume  = {46},
  number  = {4},
  pages   = {44:1--44:37},
  year    = {2014},
  doi     = {10.1145/2523813}
}

@article{castle2025largeerrors,
  title   = {A Novel Approach to Forecasting After Large Forecast Errors},
  author  = {Castle, Jennifer L. and Doornik, Jurgen A. and Hendry, David F.},
  journal = {Journal of Forecasting},
  volume  = {45},
  number  = {2},
  pages   = {837--849},
  year    = {2026},
  doi     = {10.1002/for.70062}
}

@article{welford1962note,
  title   = {Note on a Method for Calculating Corrected Sums of Squares and Products},
  author  = {Welford, B. P.},
  journal = {Technometrics},
  volume  = {4},
  number  = {3},
  pages   = {419--420},
  year    = {1962},
  doi     = {10.1080/00401706.1962.10490022}
}

@article{huber1964robust,
  title   = {Robust Estimation of a Location Parameter},
  author  = {Huber, Peter J.},
  journal = {The Annals of Mathematical Statistics},
  volume  = {35},
  number  = {1},
  pages   = {73--101},
  year    = {1964},
  doi     = {10.1214/aoms/1177703732}
}

@incollection{liu2023handling,
  title     = {Handling Concept Drift in Global Time Series Forecasting},
  author    = {Liu, Ziyi and Godahewa, Rakshitha and Bandara, Kasun and Bergmeir, Christoph},
  booktitle = {Forecasting with Artificial Intelligence: Theory and Applications},
  pages     = {163--189},
  publisher = {Palgrave Macmillan},
  address   = {Cham},
  year      = {2023},
  doi       = {10.1007/978-3-031-35879-1_7}
}

@inproceedings{derakhshan2019continuous,
  title     = {Continuous Deployment of Machine Learning Pipelines},
  author    = {Derakhshan, Behrouz and Rezaei Mahdiraji, Alireza and Rabl, Tilmann and Markl, Volker},
  booktitle = {Proceedings of the 22nd International Conference on Extending Database Technology},
  pages     = {397--408},
  year      = {2019},
  doi       = {10.5441/002/edbt.2019.35},
  publisher = {OpenProceedings.org},
  address   = {Konstanz, Germany}
}

@inproceedings{zeng2023dlinear,
  title     = {Are Transformers Effective for Time Series Forecasting?},
  author    = {Zeng, Ailing and Chen, Muxi and Zhang, Lei and Xu, Qiang},
  booktitle = {Proceedings of the AAAI Conference on Artificial Intelligence},
  volume    = {37},
  pages     = {11121--11128},
  year      = {2023},
  publisher = {AAAI Press},
  address   = {Washington, DC, USA}
}

@inproceedings{liu2024itransformer,
  title     = {iTransformer: Inverted Transformers Are Effective for Time Series Forecasting},
  author    = {Liu, Yong and Hu, Tengge and Zhang, Haoran and Wu, Haixu and Wang, Shiyu and Ma, Lintao and Long, Mingsheng},
  booktitle = {The Twelfth International Conference on Learning Representations},
  year      = {2024},
}

@inproceedings{wu2023timesnet,
  title     = {TimesNet: Temporal 2D-Variation Modeling for General Time Series Analysis},
  author    = {Wu, Haixu and Hu, Tengge and Liu, Yong and Zhou, Hang and Wang, Jianmin and Long, Mingsheng},
  booktitle = {International Conference on Learning Representations},
  year      = {2023}
}

@misc{electricity20112014,
  author       = {Trindade, Artur},
  title        = {{ElectricityLoadDiagrams20112014}},
  year         = {2015},
  howpublished = {UCI Machine Learning Repository},
  note         = {{DOI}: https://doi.org/10.24432/C58C86}
}

@inproceedings{lai2018lstnet,
  title={Modeling Long- and Short-Term Temporal 
         Patterns with Deep Neural Networks},
  author={Lai, Guokun and Chang, Wei-Cheng and 
          Yang, Yiming and Liu, Hanxiao},
  booktitle={The 41st International ACM SIGIR 
             Conference on Research \& Development 
             in Information Retrieval},
  pages={95--104},
  year={2018},
  doi={10.1145/3209978.3210006}
}

@inproceedings{zhou2021informer,
  title     = {Informer: Beyond Efficient Transformer for Long Sequence Time-Series Forecasting},
  author    = {Zhou, Haoyi and Zhang, Shanghang and Peng, Jieqi and Zhang, Shuai and Li, Jianxin and Xiong, Hui and Zhang, Wancai},
  booktitle = {Proceedings of the AAAI Conference on Artificial Intelligence},
  volume    = {35},
  pages     = {11106--11115},
  year      = {2021},
  publisher = {AAAI Press},
  address   = {Washington, DC, USA}
}

@inproceedings{wu2021autoformer,
  title     = {Autoformer: Decomposition Transformers with Auto-Correlation for Long-Term Series Forecasting},
  author    = {Wu, Haixu and Xu, Jiehui and Wang, Jianmin and Long, Mingsheng},
  booktitle = {Advances in Neural Information Processing Systems},
  volume    = {34},
  pages     = {22419--22430},
  year      = {2021},
  publisher = {Curran Associates, Inc.},
  address = {Red Hook, NY, USA}
}

@article{hoi2021online,
  title   = {Online Learning: A Comprehensive Survey},
  author  = {Hoi, Steven C. H. and Sahoo, Doyen and Lu, Jing and Zhao, Peilin},
  journal = {Neurocomputing},
  volume  = {459},
  pages   = {249--289},
  year    = {2021},
  doi     = {10.1016/j.neucom.2021.04.112}
}

@article{lu2019conceptdrift,
  title   = {Learning under Concept Drift: A Review},
  author  = {Lu, Jie and Liu, Anjin and Dong, Fan and Gu, Feng and Gama, Jo{\~a}o and Zhang, Guangquan},
  journal = {IEEE Transactions on Knowledge and Data Engineering},
  volume  = {31},
  number  = {12},
  pages   = {2346--2363},
  year    = {2019},
  doi     = {10.1109/TKDE.2018.2876857}
}

@misc{hoffman2024useful,
  title         = {Some Models Are Useful, but for How Long?: A Decision Theoretic Approach to Choosing When to Refit Large-Scale Prediction Models},
  author        = {Hoffman, Kentaro and Salerno, Stephen and Leek, Jeff and McCormick, Tyler},
  year          = {2024},
  eprint        = {2405.13926},
  archivePrefix = {arXiv},
  primaryClass  = {stat.ME},
  doi           = {10.48550/arXiv.2405.13926}
}

@article{hastings1947low,
  title={Low moments for small samples: a comparative study of order statistics},
  author={Hastings Jr, Cecil and Mosteller, Frederick and Tukey, John W and Winsor, Charles P},
  journal={The Annals of Mathematical Statistics},
  volume={18},
  number={3},
  pages={413--426},
  year={1947}
}

@inproceedings{dougherty1995supervised,
  title     = {Supervised and Unsupervised Discretization of Continuous Features},
  author    = {Dougherty, James and Kohavi, Ron and Sahami, Mehran},
  booktitle = {Proceedings of the Twelfth International Conference on Machine Learning (ICML)},
  pages     = {194--202},
  year      = {1995},
  publisher = {Morgan Kaufmann},
  doi       = {10.1016/B978-1-55860-377-6.50032-3}
}

@article{he2009learning,
  title={Learning from imbalanced data},
  author={He, Haibo and Garcia, Edwardo A},
  journal={IEEE Transactions on Knowledge and Data Engineering},
  volume={21},
  number={9},
  pages={1263--1284},
  year={2009},
  publisher={IEEE}
}

@inproceedings{dietterich2000ensemble,
  title={Ensemble Methods in Machine Learning},
  author={Dietterich, Thomas G.},
  booktitle={Multiple Classifier Systems},
  series={Lecture Notes in Computer Science},
  volume={1857},
  pages={1--15},
  year={2000},
  publisher={Springer},
  doi={10.1007/3-540-45014-9_1}
}

@book{kuncheva2004combining,
  title={Combining Pattern Classifiers: Methods and Algorithms},
  author={Kuncheva, Ludmila I.},
  year={2004},
  publisher={John Wiley \& Sons}
}

@article{wilcoxon1945test,
  title={Individual comparisons by ranking methods},
  author={Wilcoxon, Frank},
  journal={Biometrics bulletin},
  volume={1},
  number={6},
  pages={80--83},
  year={1945},
  publisher={JSTOR}
}

@article{grundy2026online,
author = {Grundy, Thomas and Killick, Rebecca and Svetunkov, Ivan},
title = {Online Detection of Forecast Model Inadequacies Using Forecast Errors},
journal = {Journal of Time Series Analysis},
volume = {47},
number = {3},
pages = {715--726},
doi = {https://doi.org/10.1111/jtsa.12843},
year = {2026}
}

@article{pearson1895note,
  title={Note on Regression and Inheritance in the Case of Two Parents},
  author={Pearson, Karl},
  journal={Proceedings of the Royal Society of London},
  volume={58},
  pages={240--242},
  year={1895},
  publisher={JSTOR}
}

\end{document}